\documentclass{article}
\usepackage[T1]{fontenc}
\usepackage[utf8]{inputenc}
\usepackage{color}
\usepackage{array}
\usepackage{booktabs}
\usepackage{bm}
\usepackage{multirow}
\usepackage{amsmath}
\usepackage{amssymb}
\usepackage{graphicx}
\usepackage[authoryear]{natbib}
\usepackage[unicode=true,
 bookmarks=false,
 breaklinks=false,pdfborder={0 0 1},backref=section,colorlinks=false]
 {hyperref}

\makeatletter

\providecommand{\tabularnewline}{\\}

\usepackage{url}
\usepackage{amsfonts}
\usepackage{nicefrac}
\usepackage{microtype}
\usepackage{bm}\usepackage{multirow}
\usepackage{array}
\usepackage{rotating}
\usepackage{float}
\usepackage{wrapfig}
\usepackage{lipsum}\DeclareMathOperator*{\argmin}{argmin}
\usepackage{lipsum}\DeclareMathOperator*{\argmax}{argmax}
\usepackage{times}
\setcitestyle{numbers}
\setcitestyle{square}
\usepackage{amsthm}\usepackage{caption}\usepackage{bm}\numberwithin{equation}{section}
\newtheorem{lemma}{\textbf{Lemma}}\usepackage{algorithmic}

\usepackage{microtype}
\usepackage{subfigure}

\usepackage[accepted]{icml2018}

\icmltitlerunning{\emph{MSplit} LBI: Realizing Feature Selection and Dense Estimation Simultaneously}

\makeatother

\begin{document}
\twocolumn[
\icmltitle{\emph{MSplit} LBI: Realizing Feature Selection and Dense Estimation Simultaneously in Few-shot and Zero-shot Learning}

\icmlsetsymbol{equal}{$\dagger$}
\icmlsetsymbol{corresp}{$\ddagger$}

\begin{icmlauthorlist}
\icmlauthor{Bo Zhao}{equal,to}
\icmlauthor{Xinwei Sun}{equal,goo}
\icmlauthor{Yanwei Fu}{corresp,ed,nd}
\icmlauthor{Yuan Yao}{corresp,fo}
\icmlauthor{Yizhou Wang}{to}
\end{icmlauthorlist}

\icmlaffiliation{to}{Nat’l Eng. Lab. for Video Technology; Key Lab. of Machine Perception (MoE); Cooperative Medianet Innovation Center, Shanghai; Sch’l of EECS, Peking University. Deepwise Inc. \\} 
\icmlaffiliation{goo}{Sch’l of Mathematical Science, Peking University. Deepwise Inc. \\}
\icmlaffiliation{ed}{Sch’l of Data Science, Fudan University.} 
\icmlaffiliation{nd}{AITrics Inc.\\}
\icmlaffiliation{fo}{Hong Kong University of Science and Technology; Peking University}

\icmlcorrespondingauthor{Yanwei Fu}{yanweifu@fudan.edu.cn}
\icmlcorrespondingauthor{Yuan Yao}{yuany@ust.hk}


\vskip 0.3in
]



\printAffiliationsAndNotice{\icmlEqualContribution} 

\begin{abstract}
It is one typical and general topic of learning a good embedding model to efficiently learn the representation coefficients between two spaces/subspaces. To solve this task, $L_{1}$ regularization is widely used for the pursuit of feature selection and avoiding overfitting, and yet the sparse estimation of features in $L_{1}$ regularization may cause the underfitting of training data. $L_{2}$ regularization is also frequently used, but it is a biased estimator. In this paper, we propose the idea that the features consist of three orthogonal parts, \emph{namely} sparse strong signals, dense weak signals and random noise, in which both strong and weak signals contribute to the fitting of data. 
To facilitate such novel decomposition, \emph{MSplit} LBI is for the first time proposed to realize feature selection and dense estimation simultaneously. 
We provide theoretical and simulational verification that our method exceeds $L_{1}$ and $L_{2}$ regularization, and extensive experimental results show that our method achieves state-of-the-art performance in the few-shot and zero-shot learning.
\end{abstract}
\section{Introduction}

This paper discusses the problem of learning representation coefficients
between two spaces/subspaces. This is one typical and general research
topic that can be used in various tasks, such as learning feature
embedding in Few-shot learning (FSL) and capturing relational
structures in Zero-Shot Learning (ZSL) (\citet{palatucci2009zero}).
In particular, FSL (\citet{fei2006one}) aims to learn new concepts
with only few training samples, while ZSL tends to learn new concepts
without any training samples. The semantic spaces such as attributes
\citet{lampert13AwAPAMI}, textual descriptions \citet{deep_0shot}
and word vectors \citet{fu2016semi} are served as the auxiliary knowledge
to assist the ZSL. This paper concerns the FSL and ZSL in transfer
learning scenario. The data in the source domain is abundant to train
the feature extractors (e.g., deep Convolutional Neural Networks (CNNs)
\citet{krizhevsky2012imagenet,simonyan2014very,InceptionNet,he2016deep});
and the data in the target are very limited to learn/fine-tune a deep
model.

The natural solutions of FSL and ZSL are to learn the linear embedding
models, which can map the image features to the label space (FSL)
(or semantic space (ZSL)). To efficiently learn such a linear model, $L_{1}$
or $L_{2}$ penalty terms are frequently applied to regularize the
weights of embedding models. In particular, the $L_{1}$ regularization
can capture the strong and sparse signals in the embedding weights,
which is also a process of feature selection. Nevertheless, the feature
selection property of $L_{1}$ penalty suffers from two problems.
1) the inaccurate estimation of strong signals if irrepresentable
condition does not hold \citet{modelselection_jmlr}; 2) the underfitting
of training data due to the ignorance of weak signals from the embedding
/ relational weights. In contrast, $L_{2}$ penalty yet does a proportional
shrinkage of feature dimension, and thus it may introduce the bias
in learning the embedding model. However, in real-world applications,
it is of equal importance to do both the feature selection and well
data-fitting. For example, in Text Classification (\citet{forman2003extensive}),
Bioinformatics (\citet{saeys2007review}) and Neuroimaging Analysis
(\citet{sun2017gsplit}), researchers need to fit the training data
well; and meantime, select a few strong signals (features) which are
comprehensible for human beings.

\begin{figure}[t]
\centering{}\includegraphics[width=1\columnwidth]{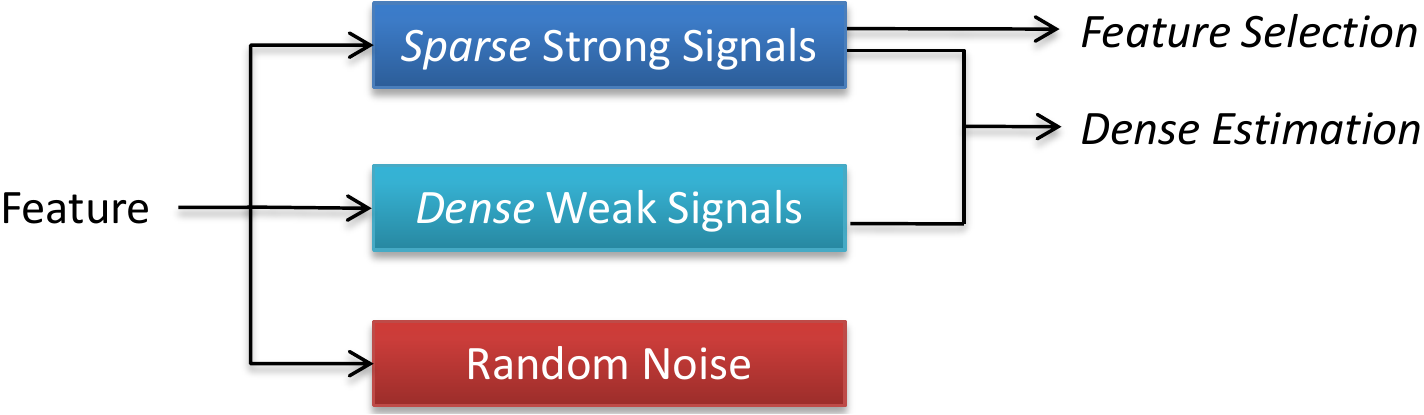}
\caption{\emph{MSplit} LBI is learned in linear embedding to decompose the
features into sparse strong signals, dense weak signals and random
noise. The sparse strong signals facilitate the feature selection.
The dense estimation can be done via the sparse strong signals and
dense weak signals.}
\label{figure:simulation}
\end{figure}

In this paper, we propose that the embedding features consist of random
noise, sparse strong signals and dense weak signals. In Sec. \ref{subsec:Multiple-Split-LBI},
the \emph{MSplit} LBI is for the first time proposed to facilitate
the decomposition of features. Particularly, in our linear embedding
models, the  {MSplit} LBI will decompose
the embedding weights into three orthogonal parts (in Sec.\ref{subsec:Decomposition-property-of}),
\emph{namely}, random noise, sparse strong signals and dense weak
signals as illustrated in Fig. \ref{figure:simulation}. The sparse
strong signals can serve the purpose of feature selection, and the
dense estimation can be done by integrating both sparse strong signals and dense weak
signals. Furthermore, we theoretically analyze the property of  {MSplit}
LBI estimator in Sec. \ref{subsec:Theoretical-Analysis-of} which
is less biased than $L_{2}-$regularization and can facilitate the
feature selection at the same time. We further show the way of using
proposed  {MSplit} LBI in FSL and ZSL tasks in Sec. \ref{subsec:Learning-by-Multiple}.
Extensive experiments had been done to validate the proposed  {MSplit}
LBI can learn better embedding models.

\noindent \textbf{Contribution}. The main contributions are several
folds: (1) We for the first time propose the idea of decomposing the
feature representation into three orthogonal elements, \emph{i.e.},
strong signals, weak signals and random noise. (2) The \emph{MSplit
LBI} algorithm is for the first time proposed to facilitate such orthogonal
decomposition and thus learn two estimators: the sparse estimator
learning strong signals and the dense estimator additionally capturing
the weak signals that also contribute to the estimation.
(3) The theoretical analysis is given in terms of the advantages
over commonly applied regularization penalties (\emph{e.g.,} $L_{1}$,
$L_{2}$ or elastic net); (4) The benefits and effectiveness of proposed
methodology are demonstrated on simulation experiments and tasks of
feature embedding learning in FSL and relational structure learning
in ZSL.

\section{Related Work}

\subsection{Feature Selection and Variable Split}

\noindent \textbf{Feature Selection.} The advantages of feature selection
can be many folds, such as avoid overfitting, or mining the correlations
between features and responses, or reducing the time complexity of
inference. Existing supervised feature selection methods can be classified
into filter methods \citet{yu2003feature}, wrapper methods \citet{kabir2010new}
and embedded methods \citet{saeys2007review} integrating the feature
selection with the classification model. Compared to filter methods,
embedded methods have the superiority that they integrate well with
classifiers. Furthermore, embedded methods are more computationally
efficient than wrapper methods. Therefore, embedded methods, e.g.
$L_{0}$ and $L_{1}$ regularization, are widely used in many real-world
applications; for instance, object recognition \citet{kavukcuoglu2010fast},
face recognition \citet{wright2009robust}, image restoration \citet{mairal2009non},
subspace clustering \citet{elhamifar2009sparse}, few-shot learning
\citet{lee2015communication} and zero-shot learning \citet{kodirov2015unsupervised}.
Because the optimization with $L_{0}$ regularization is NP hard,
$L_{1}$ regularization, which is the tightest convex relaxation of
$L_{0}$, is used for the sparsity in the most practice.

\noindent \textbf{Variable Split.} To deal with $L_{1}$ penalty and
other constrains, the operator splitting ideas are adopted by introducing
an augmented variable satisfying the sparsity (or non-negative) constraints,
such as ADMM \citet{wahlberg2012admm,boyd2011distributed}. By adopting
such schemes, the two estimators are introduced and split apart with
one being dense and the augmented one pursuing sparsity requirements.
For example in \citet{ye2011split,Splitlbi}, the variable splitting
scheme is proposed to avoid dealing with structural sparsity (such
as fused lasso or total variation penalty) directly. In addition to
the computational advantage, the \citet{sun2017gsplit} discussed
another benefit of variable splitting term that by relaxing the distance
between two estimators, the dense estimator can show a better prediction
power than the sparse one since degree of freedom to capture extra
features that can fit data better.\textcolor{red}{{} }To achieve similar
effect, one can also use ridge or elastic net models \citet{zou2005regularization}
to select more correlated features by enforcing strictly convexity
via $L_{2}$ penalty.

\subsection{Few-shot and Zero-shot Learning}

\textbf{Few-shot Learning.} The naive way to implement few-shot learning
is fine-tuning the model (trained on the source domain) on the target
domain. However, the model will easily overfit the several training
samples and hardly generalize to testing samples. The $k$-nearest
neighbor classifier is often used as the baseline in few-shot learning
\citet{koch2015siamese,santoro2016meta}. When only one sample per
target class is provided in training, i.e. $k=1$, it can be viewed
as a linear model. The Siamese neural network is proposed by \citet{koch2015siamese},
which contains twin deep feature extractors for two input images.
The $L_{1}$ component-wise distance between two feature vectors are
computed with the sigmoid activation function. \citet{snell2017prototypical}
propose the prototypical network which is combined by the deep feature
extractor and linear model. Testing images are mapped into the learned
embedding space, then classified based on a softmax over distances
to class prototypes.

\noindent \textbf{Zero-shot Learning.} Currently, most popular methods
in ZSL are linear models, as deep models may easily overfit on target
domain \citet{zhang2016learning}. $L_{1}$ and $L_{2}$ regularization
terms are frequently used in these linear models. \citet{palatucci2009zero,li2015semi}
try to learn the linear mapping from the image feature space to semantic
embedding space. $L_{2}$ regularization is utilized to avoid overfitting.
\citet{li2015semi} considers ZSL as a sparse coding problem. They
try to regress the image features use the learned dictionary with
sparse codes (semantic embeddings). $L_{1}$ regularization is utilized
to realize sparsity. In \citet{wang2016relational,zhao2017zero},
structural knowledge is learned by linearly regressing unseen semantic
embeddings on seen ones. $L_{1}$ regularization is introduced, because
they assume the connection between seen and unseen classes is sparse.

\section{Methodology}

We take the transfer learning setting. The source domain and target
domain data are denoted as $\{(\mathbf{x}_{1}^{s},{y}_{1}^{s}),...,(\mathbf{x}_{N^{s}}^{s},{y}_{N^{s}}^{s})\}$
and $\{(\mathbf{x}_{1}^{t},{y}_{1}^{t}),...,(\mathbf{x}_{N^{t}}^{t},{y}_{N^{t}}^{t})\}$
respectively. Here $\mathbf{x}_{i}^{s}\in\Re^{d}$ and $\mathbf{x}_{i}^{t}\in\Re^{d}$
indicate the visual features; and ${y}_{k}^{s}\in\mathbf{Y}^{s}$
and ${y}_{k}^{t}\in\mathbf{Y}^{t}$ are the class labels. We concern
two different settings. (1) \emph{Few-shot setting}. Only few labeled
training images are available in the target domain. (2) \emph{Zero-shot
setting.} No training data but auxiliary knowledge is available in
target domain.

Each class is embedded in the label space and expressed as $\mathbf{e}_{k}^{s}$
and $\mathbf{e}_{k}^{t}$ $\in\Re^{p}$. The label embedding of each
class is a one-hot vector in FSL, while we use auxiliary knowledge
\citet{lampert13AwAPAMI,transductiveEmbeddingJournal} to embed each
label to be a semantic vector in ZSL. Thus we can learn the linear
classification models on training data. Without loss of generality,
we first consider the linear regression of the label embeddings $\mathbf{E}=\{\mathbf{e}_{1},...,\mathbf{e}_{N}\}\in\mathbb{R}^{N\times p}$
using visual features $\mathbf{X}=\{\mathbf{x}_{1},...,\mathbf{x}_{N}\}\in\mathbb{R}^{N\times d}$.
It is formulated as
\begin{equation}
\mathbf{E}=\mathbf{X}\mathbf{B},
\end{equation}
where $\mathbf{B}\in\Re^{d\times p}$ is the linear embedding matrix.

In particular, to learn the mapping, we optimize the following formulation,
\begin{equation}
\bm{B}=\arg\min_{\bm{B}}\:\ell(\bm{B},\mathbf{X})+\lambda\Omega\left(\bm{B}\right)\label{eq:loss_function}
\end{equation}
where $\ell(\bm{B},\mathbf{X})=\left\Vert \mathbf{X}\mathbf{B}-\mathbf{E}\right\Vert ^{2}$
is the loss function over the training samples. $\Omega\left(\bm{B}\right)$
indicates the regularization term over $\mathbf{B}$. The $\lambda$
is the regularization parameter.

\subsection{Weakness of Lasso Embedding}

\noindent Various forms of regularization have been used in previous
work such as $L_{2}-$penalty \citet{fu2016semi,romera2015embarrassingly}
and $L_{1}-$penalty \citet{kodirov2015unsupervised,wang2016relational}.
Here we want to learn sparse weights $\mathbf{B}$ to capture the
strong signals in the embedding. One can apply the $L_{1}$ regularization
as,
\begin{equation}
\arg\min_{\mathbf{B}}\ell(\mathbf{B},\mathbf{X})+\lambda\sum_{j=1}^{p}\left\Vert \mathbf{B}_{(j)}\right\Vert _{1}\label{eq:lasso}
\end{equation}
where $\mathbf{B}_{(j)}$ refers the $j$-th column. Eq (\ref{eq:lasso})
turns out to be a classical Lasso formulation which can linearly regress
the sparse strong signals and set dense weak signals to be zeros.
In general, Lasso is \emph{sign consistent} if there exists a sequence\footnote{Here $\lambda_{n}$ indicates that $\lambda$ is a function of $n$.}
{ $\lambda_{n}$ such that $P\left(\hat{\mathbf{B}}\left(\lambda_{n}\right)=_{s}\lambda^{*}\right)\rightarrow1$,
as $n\rightarrow\infty$} and if Irrepresentable Condition and Beta
Min condition hold (\citet{FanLi01,sharp_lasso,zhao2006model}). Here
we define $=_{s}$ such that $\hat{\mathbf{B}}\left(\lambda\right)=_{s}\mathbf{B}^{*}$
iff $sgn\left(\hat{\mathbf{B}}\left(\lambda\right)\right)=sgn\left(\mathbf{B}^{*}\right)$
element-wise; and {$sgn\left(x\right) = 1$ if $x > 0$; $= -1$ if $x < 0$; $= 0$ otherwise.}
$\mathbf{B}^{*}$ indicate the true sparse embedding weights with
the corresponding regularizing parameter $\lambda^{*}$.

The Irrepresentable and Beta Min conditions are not easy to be satisfied
in many real-world applications. The Irrepresentable Condition implies
the low correlation between the informative and uninformative feature
dimensions. Unfortunately, the correlated variables of features, especially
in a high-dimensional space ($p>>n$), are a perennial problem for
the Lasso. Such a problem will frequently lead to systematic failures
and an inaccurate estimation of index set of strong signals. On the
other hands, The Beta Min condition requires the strong feature dimension
of non-zero coefficients should be higher than the threshold pre-specified.
Nevertheless, some feature dimensions of weak signals that are totally
ignored by Lasso, may still be very helpful in estimating the response
variables in Eq (\ref{eq:lasso}); and thus the inferior linear embedding
mapping is usually learned than the embedding learned by ridge regression.
For example, the recent neuroimaging analysis work \citet{sun2017gsplit}
showed the lesion features (strong signals) are most contributed to
identifying the disease concerned. In addition, although ``procedural
bias\textquotedbl{} features are weak signals, they can still be leveraged
to improve the prediction of the disease.

\subsection{Multiple Split LBI\label{subsec:Multiple-Split-LBI}}

This paper targets at alleviating the Irrepresentable Condition and
capturing the weak signal in Eq (\ref{eq:lasso}). The key idea is
to generalize the Split LBI algorithm (\citet{Splitlbi}) to general
loss function with response variables embedded in multiple ($p>1$)
columns ($\mathbf{E}\in\mathbb{R}^{N_{s}\times p}$). Thus, we call
it Multiple Split LBI (\emph{MSplit LBI}). Specifically, rather than
directly dealing with $\sum_{i=1}^{p}\Vert\mathbf{B}_{(j)}\Vert_{1}$
in Eq (\ref{eq:lasso}), we introduce an augmented variable $\mathbf{\Gamma}$
of the same size as $\mathbf{B}$. Here we want $\mathbf{\Gamma}$
to: (1) be enforced sparsity of each column and select the set of
strong signals (2) be close to $\mathbf{B}$ from which the distance
is controlled by the variable splitting term $\frac{1}{2\nu}\Vert\mathbf{B}-\mathbf{\Gamma}\Vert_{F}^{2}$
in the following loss function:
\begin{align}
\ell(\mathbf{B},\mathbf{\Gamma})=\ell(\mathbf{B},\mathbf{X})+\frac{1}{2\nu}\Vert\mathbf{B}-\mathbf{\Gamma}\Vert_{F}^{2} & \left(\nu>0\right)\label{eq:split-loss}
\end{align}

To pursue the sparsity requirement of $\mathbf{\Gamma}$, we utilize
Linearized Bregman Iteration (LBI) on each column of $\mathbf{B}$
and concatenate them together (please refer supplementary material
for details), which can give a sequence of estimation as a regularization
solution path, i.e. $\{\mathbf{B}_{k},\mathbf{\Gamma}_{k},\widetilde{\mathbf{B}}_{k}\}$,
\begin{subequations} \label{eq:slbi-show}
\begin{align}
\mathbf{B}_{k+1} & =\mathbf{B}_{k}-\kappa\alpha\nabla_{\mathbf{B}}\ell(\mathbf{B}_{k},\mathbf{\Gamma}_{k}),\label{eq:slbi-show-a}\\
\mathbf{Z}_{k+1} & =\mathbf{Z}_{k}-\alpha\nabla_{\mathbf{\Gamma}}\ell(\mathbf{B}_{k},\mathbf{\Gamma}_{k}),\label{eq:slbi-show-b}\\
\mathbf{\Gamma}_{k+1} & =\kappa\cdot\mathcal{S}\left(\mathbf{Z}_{k+1},1\right),\label{eq:slbi-show-c}\\
\widetilde{\mathbf{B}}_{k+1} & =\mathrm{\mathbf{P}}_{\widetilde{\mathbf{S}}_{k+1}}\mathbf{B}_{k+1}=\mathbf{B}_{k+1}\circ\left[1\{i\in\widetilde{\mathbf{S}}_{(j),k+1}\}\right]_{\{i,j\}}\label{eq:slbi-show-d}
\end{align}
\end{subequations}
\noindent where $\mathbf{Z}_{0}=\mathbf{\Gamma}_{0}=\widetilde{\mathbf{B}}_{0}=\mathbf{0}\in\mathbb{R}^{d\times p}$,
$\widetilde{\mathbf{S}}_{k}=\mathrm{supp}(\mathbf{\Gamma}_{k})$ and
\[
\mathcal{S}\left(\mathbf{Z},\lambda\right)=\mathrm{sign}(\mathbf{Z})\cdot\max\left(|\mathbf{Z}|-\lambda,\ 0\right)\ (\lambda\geq0).
\]
By implementing the soft-thresholding $\mathcal{S}\left(\mathbf{Z},1\right)$
in~Eq(\ref{eq:slbi-show-c}), the LBI returns a path of sparse estimators
$\mathbf{\Gamma}_{k}$ with different sparsity levels at each iteration.
The parameter $t_{k}=k\alpha$ is the regularization parameter which
plays a similar role with $\lambda$ in~Eq (\ref{eq:lasso}). In
real applications, it can be determined via cross validation (please
refer to \citet{bregman} and therein). The parameter $\kappa$ is
the damping factor. The larger value of $\kappa$ can de-bias estimators
however at the sacrifice of computational efficiency. Parameter $\alpha$
is the step size, which should satisfy $\kappa\alpha\leq\nu/\kappa(2+\nu\Lambda_{H})$\footnote{Here $\Lambda(\cdot)$ denotes the largest singular value of a matrix
and $H$ denotes the Hessian matrix of $\ell(\mathbf{B})$.} (\citet{Splitlbi}) to ensure the statistical property. Parameter
$\nu$ in~Eq~(\ref{eq:split-loss}) controls the distance between
$\mathbf{B}$ and $\tilde{\mathbf{B}}$. Such two estimators are tending
to be close with smaller value of $\nu$.

\subsection{Decomposition property of MSplit LBI\label{subsec:Decomposition-property-of}}

The proposed MSplit LBI has several advantages. Most importantly,
it has the decomposition property. Specifically, the path of dense
estimators $\mathbf{B}_{k}$ computed by Eq (\ref{eq:slbi-show-a}–\ref{eq:slbi-show-d})
has the following orthogonal decomposition of elements,
\begin{align}
\mathbf{B}_{k}={\mathrm{Signal_{strong}}\oplus\mathrm{Signal_{weak}}\oplus\mathrm{Random\ Noise}}\label{eq:decomposition}
\end{align}
The strong signals are captured by the projection of $\mathbf{B}_{k}$
to the subspace of support set of $\mathbf{\Gamma}_{k}$ (Eq (\ref{eq:slbi-show-d})), {i.e. $\widetilde{\mathbf{B}}_{k}$}. {Hence $\mathbf{B}_{k}$ shares the same value of strong signals with $\widetilde{\mathbf{B}}_{k}$}. The remainder of such projection is heavily influenced by weak signals,
which are captured by non-zero elements of $\mathbf{B}_{k}$ with
comparably large magnitude, while others with tiny values are regarded
as random noise. Hence, the algorithm gives a path of two estimators:
$\mathbf{B}_{k}$ and $\widetilde{\mathbf{B}}_{k}$. {Thus, our goal includes two folds:} (1) Use $\widetilde{\mathbf{B}}_{k}$ to select the interpretable strong signals; (2) use $\mathbf{B}_{k}$ for prediction
 since it can additionally leverage weak signals for better fitness of data.

The capture of weak evidences are influenced by parameter $\nu$ and
$t_{k}$. Note that with larger value of $\nu$, the $\mathbf{B}$
has more degree of freedom to capture weak signal with less constraint
between $\mathbf{B}$ and $\widetilde{\mathbf{B}}$, and vise-versa. Besides,
it's the trade-off between (1) model selection consistency and (2)
prediction task. On one hand, the irrepresentable condition is more
easier to satisfy with larger value of $\nu$ and On the other hand,
it will lower the signal-to-noise ratio and hence deteriorate the
estimation of the parameter.

For the regularization parameter $t_{k}$, note that as the algorithm
iterates, it tends to give $\widetilde{\mathbf{B}}_{k}$ with less sparsity
levels and $\lim_{k\to\infty}\Vert \widetilde{\mathbf{B}}_{k}-\mathbf{B}_{k}\Vert_{F}^{2}\to0$.
In such case, the estimation of strong signals are inaccurate and
$\mathbf{B}$ has not degree of freedom to capture weak signals.

Compared against Lasso-type penalty, our MSplit LBI generally
has more advantages, beside of its simpler iterative scheme: (1) MSplit
LBI can capture weak signals which are ignored by $\ell_{1}$ penalty
due to the Beta Min condition. (2) According to Theorem 1 in \citet{Splitlbi},
the irrepresentable condition is more easier to be met when $\nu$
is large enough, leading to more robust model selection consistency.
(3) Combined with the less bias property of LBI, the estimation of
strong signal is more accurate than Lasso as discussed in next subsection.

\subsection{Theoretical Analysis of MSplit LBI \label{subsec:Theoretical-Analysis-of}}

\noindent \textbf{Bias Vs. Unbiased.} Although the ridge-type penalty
and elastic net can weaken the irrepresentable condition by de-correlating
column-wise correlation, the regularization parameter will introduce
bias during the estimation of strong signals. In contrast, our MSplit
LBI is unbiased estimator for strong signals and for weak signals
when $\nu\to\infty$. This section introduces two lemmas comparing
the differences.

To see this, the following lemma describes the biased estimator given
by ridge regression and elastic net under the simplest case. {In the following, we use
($\beta$, $\tilde{\beta}$) to denote the vector notation of ($\mathbf{B}$, $\mathbf{\widetilde{B}}$).}

\begin{lemma} \label{lemma:bias-ridge-elastic} Assume $y=\beta^{\star}+\varepsilon$
where $\varepsilon$ has independent identically distributed components,
each of which has a sub-Gaussian distribution with 0 mean. $\mathbf{S}=\{i:\beta_{i}^{\star}\gtrsim\sqrt{\frac{s\log{p}}{n}}\}$,
the ridge estimator and the elastic net estimators
\begin{align}
\beta_{S}^{ridge} & =\arg\min\frac{1}{2}\Vert y-\beta\Vert_{2}^{2}+\frac{\lambda_{\ell_{2}}}{2}\Vert\beta\Vert_{2}^{2}\label{eq:estimators}
\end{align}
\begin{align}
\beta_{S}^{elastic} & =\arg\min\frac{1}{2}\Vert y-\beta\Vert_{2}^{2}+\frac{\lambda_{\ell_{2}}}{2}\Vert\beta\Vert_{2}^{2}+\lambda_{\ell_{1}}\Vert\beta\Vert_{1}
\end{align}
we have
\begin{align}
\mathrm{E}(\beta_{\mathbf{S}}^{ridge}) & =\frac{1}{1+\lambda_{\ell_{2}}}\beta_{\mathbf{S}}^{\star}\label{eq:bias-ridge}\\
\mathrm{E}(\beta_{\mathbf{S}}^{elastic}) & =\frac{\beta_{\mathbf{S}}^{\star}}{1+\lambda_{\ell_{2}}}+\frac{1}{1+\lambda_{\ell_{2}}}\mathbf{E}_{\varepsilon_{\mathbf{S}}\leq-\beta_{\mathbf{S}}^{\star}-\lambda_{\ell_{1}}}(\varepsilon_{\mathbf{S}}+\lambda_{\ell_{1}})\nonumber \\
 & \frac{\beta_{\mathbf{S}}^{\star}}{1+\lambda_{\ell_{2}}}P(-\beta_{\mathbf{S}}^{\star}-\lambda_{\ell_{1}}\leq\varepsilon_{\mathbf{S}}\leq-\beta_{\mathbf{S}}^{\star}+\lambda_{\ell_{1}})\nonumber \\
 & +\frac{1}{1+\lambda_{\ell_{2}}}\mathbf{E}_{\varepsilon_{\mathbf{S}}\geq\lambda_{\ell_{1}}-\beta_{\mathbf{S}}^{\star}}(\varepsilon_{\mathbf{S}}-\lambda_{\ell_{1}})\label{eq:bias-elastic}
\end{align}
\end{lemma}

When $\kappa\to\infty$, $\alpha\to0$, the~\ref{eq:slbi-show-a} to~\ref{eq:slbi-show-c} with $\{\mathbf{B},\mathbf{\Gamma}\}$
replaced with $\{\beta,\gamma\}$ converges to 
 \begin{subequations}
\label{eq:slbi-converge}
\begin{align}
0 & =-\nabla_{\beta}\ell(\beta_{t},\gamma_{t}),\label{eq:slbi-converge-a}\\
\dot{\rho}_{\gamma_{t}} & =-\nabla_{\gamma_{\widetilde{\mathbf{S}}_{t}^{c}}}\ell(\beta_{t},\gamma_{t}),\label{eq:slbi-converge-b}\\
\rho_{\gamma_{t}} & \in\partial\Vert\gamma_{t}\Vert_{1},\label{eq:slbi-converge-c}
\end{align}
\end{subequations} Then the following lemma states that under the
case defined in lemma~\ref{lemma:bias-ridge-elastic}, we can give
more accurate estimation of $\beta_{S}^{\star}$, and also a slightly
biased estimation of $\beta_{T}^{\star}$:\\
 \begin{lemma} Under the same setting defined in lemma~\ref{lemma:bias-ridge-elastic},
if there exists $\bar{t}$ such that $\widetilde{\mathbf{S}}_{t}=\mathbf{S}$,
then $\beta_{\bar{t}}$ in~\ref{eq:slbi-converge} satisfies
\begin{align}
\beta_{\mathbf{S},\bar{t}}=\beta_{\mathbf{S}}^{\star}+\varepsilon_{\mathbf{S}},\ \beta_{\mathbf{S}^{c},\bar{t}}=\frac{\nu}{1+\nu}\beta_{\mathbf{S}^{c}}^{\star}+\frac{\nu}{1+\nu}\varepsilon_{\mathbf{S}^{c}}\label{eq:slbi-estimate}
\end{align}
and therefore
\begin{align}
\mathbf{E}(\beta_{\mathbf{S},\bar{t}})=\beta_{\mathbf{S}}^{\star},\ \mathbf{E}(\beta_{\mathbf{S}^{c},\bar{t}})=\frac{\nu}{1+\nu}\beta_{\mathbf{S}^{c}}^{\star}\label{eq:slbi-estimate-expect}
\end{align}
\label{lemma:slbi} \end{lemma} Hence, for strong signals, when $\nu$
is large enough, not only the model selection consistency is easier to satisfy compared to $\ell_{1}$, but also the estimation of
them are bias-free while $\ell_{1},\ell_{2}$ and elastic net are
with biases. Moreover, $\beta_{\bar{t}}$ can also capture weak signals
in $\mathbf{S}^{c}$ with bias dependent on $\nu$. According to~\ref{eq:slbi-estimate},
larger $\nu$ can give less bias estimation ($\frac{1}{1+\nu}\beta_{\mathbf{S}^{c}}^{\star}$)
at the sacrifice of more noise introduced ($\frac{\nu}{1+\nu}\varepsilon_{\mathbf{S}^{c}}$).
Note that the lemma~\ref{lemma:bias-ridge-elastic} and~\ref{lemma:slbi}
are given under $X=I$ in linear model, the more general cases are
left to appendix.

\subsection{Learning by Multiple Split LBI\label{subsec:Learning-by-Multiple}}

\label{sec:learningbyLBI} As mentioned in previous sections, the
MSplit LBI essentially has the advantage of extracting strong signals
and weak signals, which can efficiently learn the embedding in few-shot
and zero-shot learning scenarios. In these two tasks, the strong signals
correspond to the good sparse embedding, while the weak signals can
capture the weak evidences which are also useful to train the embedding.

\noindent \textbf{Few-shot Learning.} Our model can be directly used
to solve this task. We firstly use the source domain to learn the
feature extractor (\emph{i.e}. deep CNNs). Then, image features from
target domain are extracted using the trained CNNs. As few labeled
training data in target domain are provided, we use the MSplit LBI,
\emph{i.e.} Eq. (\ref{eq:split-loss} and \ref{eq:slbi-show}), to
learn the linear embedding $\mathbf{B}$ from image features $\mathbf{X}$
to (one-hot) label embeddings $\mathbf{E}$. Here, the label embedding
of each training datum is a one-hot vector with 1 on the position corresponding
to the label, while the values on other positions are 0. With the
learned embedding $\mathbf{B}$, a testing image is first embedded
to the label embedding space $\hat{\mathbf{e}}_{i}^{t}=\mathbf{x}_{i}^{t}\mathbf{B}$,
then labeled as the one with maximum value $\hat{y}_{i}^{t}=\argmax_{k}\hat{\mathbf{e}}_{i(k)}^{t}$.
The element $\hat{\mathbf{e}}_{i(k)}^{t}$ denotes the $k$th element
in the vector $\hat{\mathbf{e}}_{i}^{t}$.\textbf{ }

\noindent \textbf{Zero-shot Learning.} On this task, our method is
based on the structural knowledge transfer. Specifically, the structure
among classes is learned in the semantic label embedding space by
linearly regressing the label embeddings of target domain classes
($\mathbf{E}^{t}$) on source target domain classes ($\mathbf{E}^{s}$).
The Eq. (\ref{eq:split-loss}) is adapted to be $\mathbf{E}^{t}=\mathbf{E}^{s}\mathbf{B}$.
Similar as \citet{snell2017prototypical} , we use the prototype to
represent each class and implement nearest neighbour classification
in the image feature space. The prototype of each source domain class
is calculate as the mean vector of all samples in the class, i.e.
$\mathbf{f}_{k}^{s}=\frac{1}{N^s_k}\sum_{i}^{N^s_k}{\mathbf{x}_{i}^{s}}\;\;s.t.\;y_{i}^{s}=k$.
${N^s_k}$ denotes the number of training samples from the $k$th seen class.
Then the learned structure ($\mathbf{B}$) is transferred to the image
feature space for synthesizing the prototypes of all target domain
classes $\hat{\mathbf{F}}^{t}=\mathbf{F}^{s}\mathbf{B}$, where $\mathbf{F}^{s}=\{\mathbf{f}_{1}^{s},...,\mathbf{f}_{K^{s}}^{s}\}$
and $\hat{\mathbf{F}}^{t}=\{\hat{\mathbf{f}}_{1}^{t},...,\hat{\mathbf{f}}_{K^{t}}^{t}\}$
denote all prototypes in {source and target domain} respectively.
A testing image is classified based on the distance to these synthesized
prototypes $\hat{y}_{i}^{t}=\argmin_{k}{\|\mathbf{x}_{i}^{t}-\hat{\mathbf{f}}_{k}^{t}\|_{F}}$.
In the experiments, we will illustrate the learned strong
and weak signals in our model.

\section{Experiments}

In this section, we conduct three parts of experiments. First, the simulation
experiments are conducted to statistically validate the advantages of our
MSplit LBI over Lasso ($L_{1}-$penalty), Ridge Regression ($L_{2}-$penalty)
and Elastic Net. Furthermore, we have the experiments on zero-shot
and few-shot learning to illustrate the effectiveness of our model.
Finally, some visible evidences about the captured strong and weak
signals are shown in the Sec. \ref{sec:exp-vis}.

\subsection{Simulation Experiments}

In this section, we conduct a simulation experiment. We set $N=100$,
$p=80$ and generate $X\in\mathbb{R}^{N\times d}$ denoting $N$ $i.i.d$
samples from $\mathcal{N}(0,\Sigma)$ with $\Sigma_{i,j}=1$ for $(i=j)$
and $=\sigma$ for $i\neq j$. We consider four settings in which
$\sigma$ increases from 0.2 to 0.8 with space 0.2. Then we generate
$y=X\beta^{\star}+\varepsilon$ with $\beta_{i}^{\star}=2$ if $1\leq i\leq5$;
$=0.2$ if $6\leq i\leq40$; $=0$ otherwise and $\varepsilon\sim\mathcal{N}(0,0.5\cdot I_{N})$.

\begin{table}[]
\begin{centering}
\small
\begin{tabular}{ccccc}
\hline
 & {\small{}{}{}{}$\sigma=0.2$}  & {\small{}{}{}{}$\sigma=0.4$}  \tabularnewline
\hline
\hline
{\small{}{}{}{}MLE}  & {\small{}{}{}{}0.2368 $\pm$ 0.0449}  & {\small{}{}{}{}0.2739 $\pm$ 0.0516 }  \tabularnewline
\hline
{\small{}{}{}{}Ridge}  & {\small{}{}{}{}0.2057 $\pm$ 0.0388}  & {\small{}{}{}{}0.2324 $\pm$ 0.0403} \tabularnewline
\hline
{\small{}{}{}{}Elastic Net}  & {\small{}{}{}{}0.1295 $\pm$ 0.0167}  & {\small{}{}{}{}0.1359 $\pm$ 0.0178}  \tabularnewline
\hline
{\small{}{}{}{}Lasso}  & {\small{}{}{}{}0.1323 $\pm$ 0.0166}  & {\small{}{}{}{}0.1418 $\pm$ 0.0187}  \tabularnewline
\hline
{\small{}{}{}{}MSplit LBI ($\tilde{\beta}$)}  & {\small{}{}{}{}0.1463 $\pm$ 0.0241}  & {\small{}{}{}{}0.1714 $\pm$ 0.0255}  \tabularnewline
\hline
{\small{}{}{}{}MSplit LBI ($\beta$){}{}}  & \textbf{\small{}{}{}{}0.1238 $\pm$ 0.0112}{\small{}{}{}{}}  & \textbf{\small{}{}{}{}0.1312 $\pm$ 0.0117}{\small{}{}{}{}}  \tabularnewline
\midrule
& {\small{}{}{}{}$\sigma=0.6$}  & {\small{}{}{}{}$\sigma=0.8$ } \tabularnewline
\hline
\hline
{\small{}{}{}{}MLE} & {\small{}{}{}{}0.3358 $\pm$ 0.0630}  & {\small{}{}{}{}0.4751  $\pm$ 0.0891}  \tabularnewline
\hline
{\small{}{}{}{}Ridge} & {\small{}{}{}{}0.2723 $\pm$ 0.0423}  & {\small{}{}{}{}0.3479  $\pm$ 0.0455}  \tabularnewline
\hline
{\small{}{}{}{}Elastic Net} & {\small{}{}{}{}0.1643 $\pm$ 0.0241}  & {\small{}{}{}{}0.2265  $\pm$ 0.0263}\tabularnewline
\hline
{\small{}{}{}{}Lasso} & {\small{}{}{}{}0.1777 $\pm$ 0.0279}  & {\small{}{}{}{}0.2369  $\pm$ 0.0254} \tabularnewline
\hline
{\small{}{}{}{}MSplit LBI ($\tilde{\beta}$)} & {\small{}{}{}{}0.2016 $\pm$ 0.0206}  & {\small{}{}{}{}0.2421  $\pm$ 0.0159} \tabularnewline
\hline
{\small{}{}{}{}MSplit LBI ($\beta$){}{}} & \textbf{\small{}{}{}{}0.1461 $\pm$ 0.0120}{\small{}{}{}{}}  & \textbf{\small{}{}{}{}0.1749 $\pm$ 0.0127}{\small{}{}{}{} } \tabularnewline
\hline
\end{tabular}
\par\end{centering}
\caption{The estimation error of $\Vert\hat{\beta}-\beta^{\star}\Vert_{2} / \Vert\beta^{\star}\Vert_{2}$}
\label{table:acc} {\small{}{\small{}\vspace{-0.1in}}}{\small \par}
\end{table}

\begin{figure}[h!]
\centering{}\centering{}\includegraphics[width=0.9\columnwidth]{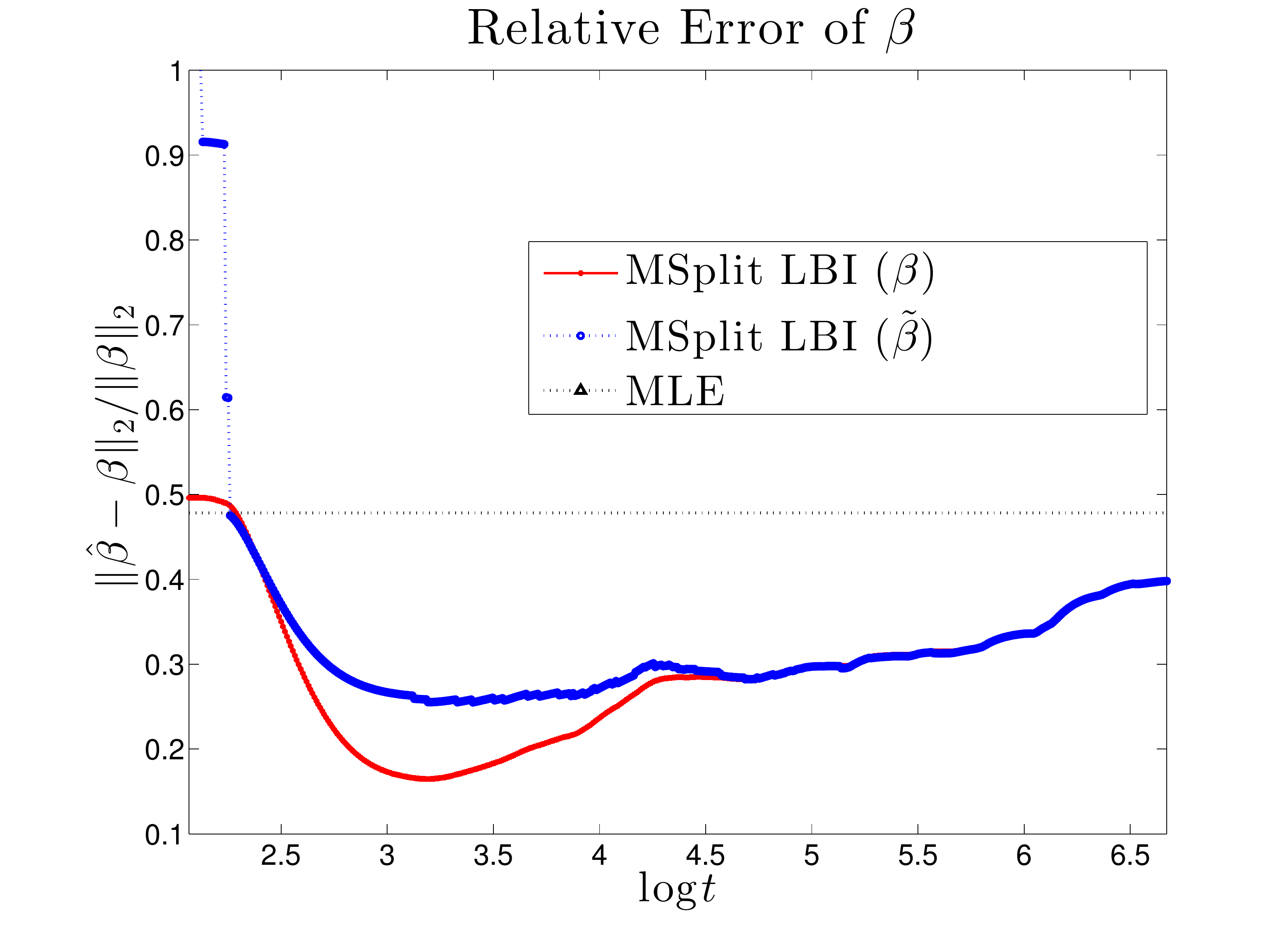}
\caption{The comparative error of $\{\beta_{k},\tilde{\beta}_{k}\}$ in the regularization
solution of MSplit LBI. Red curve represents the $\beta_{k}$ (the
dense estimator that can capture both strong and weak evidences);
blue curve represents the $\tilde{\beta}_{k}$ (the sparse estimator
that captures strong evidences); black dot line represents the estimation
of MLE.}
{\small{}{\small{}\vspace{-0.1in}}}\label{figure:simulation-1}
\end{figure}

We compare Maximum Likelihood Estimator (MLE), Lasso, Ridge, Elastic
Net and two estimators, $\beta_{k}$ and $\tilde{\beta}_{k}$ (counterparts
of $\mathbf{B}_{k}$ and $\widetilde{\mathbf{B}}_{k}$ in~\ref{eq:slbi-show})
in MSplit LBI. For $\lambda$ in Lasso, Ridge and Elastic Net, it
ranges from $\{0,0.002\cdot\lfloor\lambda_{\max}/(500-1)\rfloor,0.004\cdot\lfloor\lambda_{\max}/(500-1)\rfloor,...,\lambda_{\max}\}$,
which $\lambda_{\max}$ (which we take 5 here) is large enough in
our settings to be ensured greater than $\max_{i}\{\frac{X_{i}^{T}y}{N}\}$.
For mixture parameter $\alpha$ in Elastic Net, it's optimized from
$\{0,0.05,0.1,...,1\}$. For MSplit LBI, we set $\kappa=5$, $\alpha=\frac{\kappa}{(2+\nu\Vert X^{\star}X\Vert_{2})}$.
The parameter $\nu$ varies with $\sigma$, it is set to 3 if $\sigma=0.2$,
5 if $\sigma=0.4$ 3, 7 if $\sigma=0.6$ and 10 if $\sigma=0.4$.
In each setting ($\sigma$), we simulated 20 times and in each time,
we recorded the minimum comparative error of $\beta^{\star}$ optimized
from gird of parameters of each method.

As shown in Tab.~\ref{table:acc}, the $\beta$ of MSplit LBI outperforms
others in all cases. Note that $\beta$ is superior than $\tilde{\beta}$
since the former can capture weak evidences. {Besides, the advantage over lasso is more obvious when $\sigma$ grows}. Particularly,
when $\sigma=0.8$, the irrepresentable condition is hard to be satisfied
for Lasso while easier for MSplit LBI when $\nu$ is large enough.

The Fig.~\ref{figure:simulation-1} shows the curve of comparative
error of $\{\beta_{k},\tilde{\beta}_{k}\}$ in the regularization
solution. From the start when $k=1\thinspace(t_{k}=0)$, $\tilde{\beta}=0$
and $\beta$ is the solution of ridge regression with $\lambda=\frac{1}{2\nu}$.
As $t_{k}$ evolves, more strong signals are selected and $\beta_{k}$
is more similar to $\tilde{\beta}_{k}$ in strong evidences. At some
point (near 3.2), the $\beta_{k}$ can not only capture strong evidences
and also can capture weak evidences to fit data better. When $t_{k}$
continues to grow, the $\Vert\beta_{k}-\tilde{\beta}_{k}\Vert_{2}$
gets smaller (blue curve ($\tilde{\beta}_{k}$) and red curve ($\beta_{k}$)
converges together).

\subsection{Zero-shot Learning}

\noindent
\begin{table}
\small
\centering{}%
\begin{tabular}{ccccc}
\toprule
\multirow{2}{*}{{\small{}Methods} } & \multirow{2}{*}{{\small{}AwA} } & \multirow{2}{*}{{\small{}CUB} } & \multicolumn{2}{c}{{\small{}{}{}{}{}ImageNet }}\tabularnewline
\cmidrule{4-5}
 &  &  & {\small{}{}{}{}{}Top-1}  & {\small{}Top-5}\tabularnewline
\midrule
\midrule
{\small{}{}{}{}{}DAP/IAP}  & {\small{}{}{}{}{}41.4/42.2} & {\small{}{}{}{}{}-}  & –  & –\tabularnewline
\midrule
{\small{}{}{}{}{}SJE}  & {\small{}{}{}{}{}66.7}  & {\small{}{}{}{}{}50.1}  & –  & –\tabularnewline
\midrule
{\small{}{}{}{}{}SC\_struct}  & {\small{}{}{}{}{}72.9}  & {\small{}{}{}{}{}54.7}  & –  & –\tabularnewline
\midrule
{\small{}{}{}{}{}LatEm}  & {\small{}{}{}{}{}76.1}  & {\small{}{}{}{}{}47.4}  & –  & –\tabularnewline
\midrule
{\small{}{}{}{}{}SS-Voc}  & {\small{}{}{}{}{}78.3}  & {\small{}{}{}{}{}-}  & {\small{}{}{}{}{}{9.50}$^{\ddagger}$}  & {\small{}16.80$^{\ddagger}$ }\tabularnewline
\midrule
{\small{}{}{}{}{}ConSE}  & {\small{}{}{}{}{}-}  & {\small{}{}{}{}{}-}  & {\small{}{}{}{}{}7.80}  & {\small{}15.50 }\tabularnewline
\midrule
{\small{}{}{}{}{}DeViSE}  & {\small{}{}{}{}{}-}  & {\small{}{}{}{}{}-}  & {\small{}{}{}{}{}5.20}  & {\small{}12.80 }\tabularnewline
\midrule
{\small{}{}{}{}{}JLSE}  & {\small{}{}{}{}{}80.46}  & {\small{}{}{}{}{}42.11}  & –  & –\tabularnewline
\midrule
{\small{}{}{}{}{}MFMR-joint}  & {\small{}{}{}{}{}83.5}  & {\small{}{}{}{}{}53.6}  & –  & –\tabularnewline
\midrule
\midrule
{\small{}{}{}{}{}ESZSL}  & {\small{}{}{}{}{}79.53}  & {\small{}{}{}{}{}51.90}  & –  & –\tabularnewline
\midrule
{\small{}{}{}{}{}RKT}  & {\small{}{}{}{}{}81.41}  & {\small{}{}{}{}{}55.59}  & –  & –\tabularnewline
\midrule
{\small{}{}{}{}{} Lasso}  & {\small{}{}{}{}{} 83.19}  & {\small{}{}{}{}{} 56.00}  & {\small{}{}{}{}{}8.30}  & {\small{}{}{}{}{}18.72 }\tabularnewline
\midrule
{\small{}{}{}{}{} Ridge}  & {\small{}{}{}{}{} 83.72}  & {\small{}{}{}{}{} 51.00}  & {\small{}{}{}{}{} 6.47}  & {\small{}{}{}{}{} 16.80 }\tabularnewline
\midrule
{\small{}{}{}{}{}MSplit LBI ($\tilde{\beta}$)}  & {\small{}{}{}{}{}{84.58}}  & {\small{}{}{}{}{}{56.62}}  & {\small{}{}{}{}{}7.98}  & {\small{}17.83 }\tabularnewline
\midrule
{\small{}{}{}{}{}MSplit LBI ($\beta$)}  & \textbf{\small{}{}{}{}{}85.34}{\small{}{}{}{}{}}  & \textbf{\small{}{}{}{}{}57.52}{\small{}{}{}{}{}}  & \textbf{\small{}{}{}{}{}8.35}{\small{}{}{}{}{}}  & \textbf{\small{}18.76}{\small{} }\tabularnewline
\bottomrule
\end{tabular}{\small{}\caption{{\small{}Comparison to the state-of-the-art (\%).  $^{\ddagger}$
means that extra vocabulary information (nearly 310k word vectors)
is used. ESZSL, RKT, Lasso and Ridge are implemented using the same image features, while others are reported results.} }
{\small{}\vspace{-0.1in}}\label{tab:comparison}}
\end{table}

\noindent \textbf{Datasets.} We evaluate our method on three datasets
– Animals with Attributes (AwA) (\citet{lampert13AwAPAMI}), Caltech-UCSD
Birds-200-2011 (CUB) (\citet{WahCUB_200_2011}) and ImageNet 2012/2010
(\citet{deng2009imagenet}. These three datasets are widely used for
evaluating ZSL algorithms. AwA is a coarse-grained dataset which contains
images of 50 kinds of common animals. 85 binary and continuous attributes
are provided. As the standard split (\citet{lampert13AwAPAMI}), 10
classes are used as the target domain (unseen) classes with the rest
source domain (seen) classes. CUB is a fine-grained dataset that contains
200 kinds of birds. 312-dim continuous-valued attributes are provided.
As in \citet{akata2013label}, 50 classes are used as the target domain
(unseen) classes. The rest 150 classes are source domain (seen) classes.
ImageNet 2012/2010 is a large-scale image dataset. No attributes are
provided in this dataset. Following the setting in \citet{fu2016semi},
we use 1,000 classes in ImageNet 2012 as source domain classes.
360 classes in ImageNet 2010 which do not exist in ImageNet 2012 serve
as target domain classes.

\noindent \textbf{Competitors and Settings.} Our embedding model is
compared against the state-of-the-art methods, including DAP/IAP (\citet{lampert2014attribute}),
SJE (\citet{akata2015evaluation}), SC\_struct (\citet{changpinyo2016synthesized}),
LatEm (\citet{xian2016latent}), LEESD (\citet{dinglow}), SS-Voc
(\citet{fu2016semi}), DCL (\citet{guo2017zero}), JLSE (\citet{zhang2016zero}),
ESZSL (\citet{romera2015embarrassingly}), RKT (\citet{wang2016relational})
and MFMR-joint (\citet{xu2017matrix}). We also compare two baseline
methods, \emph{i.e.} those using Lasso and Ridge Regression (\citet{palatucci2009zero})
as the embedding methods for ZSL. For AwA and ImageNet, we use VGG-19
models pre-trained on ImageNet2012 as feature extractor. For the fine-grained
CUB dataset, we concatenate the GoogLeNet and ResNet features both
pre-trained on ImageNet 2012 dataset. We compare all ZSL methods under
the inductive settings. In other word, we donot have the features
of testing instances in the training stage, not as the transductive
setting in \citet{fu2015transductive}.

\noindent \textbf{Features.} The visual representations of images
(i.e. visual features) are very important in ZSL. Here, to make
the ZSL results more comparable, we implement the ESZSL, RKT, Lasso
and Ridge,  {MSplit} LBI by using the same image features on each
dataset. For the rest results,
we report the best results in their corresponding papers.

\noindent \textbf{Results.} We compare the performance of different
ZSL methods in Tab. \ref{tab:comparison}. It is obvious that our
method MSplit LBI ($\beta$) achieves very competitive results on
three datasets. In particular, on AwA and CUB datasets, our model
can achieve the highest accuracy of $85.34\%$ and $57.52\%$. On
ImageNet dataset, our results achieve $8.35\%$ Top-1 and $18.76\%$
Top-5 accuracy. Note that (1) though our Top-1 result is slightly
worse than $9.50\%$ Top-1 accuracy reported in \citet{fu2016semi},
the results of SS-Voc are using large scale word vectors to help inform
the learning process. (2) We can find that, except on AwA, Lasso method
always performs better than ridge method. This shows the importance
of learning the good sparse strong signals. In contrast, on small-scale
datasets (AwA and CUB), our MSplit LBI ($\beta$) obviously outperforms
the Lasso method by a clear margin – $2.15\%$ and $1.52\%$ respectively.
(3) Our sparse model – MSplit LBI ($\tilde{\beta}$), also achieves
comparable results on all these datasets and yet slightly lower results
than dense estimation model – MSplit LBI (${\beta}$). For example,
the results of MSplit LBI ($\tilde{\beta}$) is around $1\%$ lower
than those of MSplit LBI (${\beta}$) on these dataset. This performance
gap verifies that those dense weak signals of embedding also contribute
to the learning of linear embedding, and again thanks to the decomposition
ability of our MSplit LBI of being able to capture both the strong
and weak signals.

\begin{table*}
\centering{}%
\small
\begin{tabular}{cllllll}
\toprule
\multicolumn{1}{c}{{\small{}{}{}{Unseen}}} & \multicolumn{3}{c}{{\small{}{}{}{Strong Signals}}} & \multicolumn{3}{c}{{\small{}{}{}{Weak Signals}}}\tabularnewline
\midrule
{\small{}{}{}{Pig }}  & {\small{}{}{}{}}%
\begin{tabular}{@{}l@{}}
{\small{}{}{}{0.4267}}\tabularnewline
{\small{}{}{}{cow}}\tabularnewline
\end{tabular} & {\small{}{}{}{}}%
\begin{tabular}{@{}l@{}}
{\small{}{}{}{0.1837}}\tabularnewline
{\small{}{}{}{rhinoceros}}\tabularnewline
\end{tabular} & {\small{}{}{}{}}%
\begin{tabular}{@{}l@{}}
{\small{}{}{}{0.1375}}\tabularnewline
{\small{}{}{}{ox}}\tabularnewline
\end{tabular} & \multicolumn{1}{l}{{\small{}{}{}{}}%
\begin{tabular}{@{}l@{}}
\textit{\small{}{}{}0.0182}\tabularnewline
\textit{\small{}{}{}hamster}\tabularnewline
\end{tabular}} & \multicolumn{1}{l}{{\small{}{}{}{}}%
\begin{tabular}{@{}l@{}}
\textit{\small{}{}{}0.0152}\tabularnewline
\textit{\small{}{}{}skunk}\tabularnewline
\end{tabular}} & {\small{}{}{}{}}%
\begin{tabular}{@{}l@{}}
\textit{\small{}{}{}0.0151}\tabularnewline
\textit{\small{}{}{}chihuahua}\tabularnewline
\end{tabular}{\small{}{}{}{ }}\tabularnewline
\midrule
{\small{}{}{}{Hippopotamus }}  & {\small{}{}{}{}}%
\begin{tabular}{@{}l@{}}
{\small{}{}{}{0.4002}}\tabularnewline
{\small{}{}{}{rhinoceros}}\tabularnewline
\end{tabular} & {\small{}{}{}{}}%
\begin{tabular}{@{}l@{}}
{\small{}{}{}{0.3482}}\tabularnewline
{\small{}{}{}{elephant}}\tabularnewline
\end{tabular} & {\small{}{}{}{}}%
\begin{tabular}{@{}l@{}}
{\small{}{}{}{0.2214}}\tabularnewline
{\small{}{}{}{blue whale}}\tabularnewline
\end{tabular} & \multicolumn{1}{l}{{\small{}{}{}{}}%
\begin{tabular}{@{}l@{}}
\textit{\small{}{}{}-0.0163}\tabularnewline
\textit{\small{}{}{}antelope}\tabularnewline
\end{tabular}} & \multicolumn{1}{l}{{\small{}{}{}{}}%
\begin{tabular}{@{}l@{}}
\textit{\small{}{}{}-0.0136}\tabularnewline
\textit{\small{}{}{}cow}\tabularnewline
\end{tabular}} & {\small{}{}{}{}}%
\begin{tabular}{@{}l@{}}
\textit{\small{}{}{}0.0132}\tabularnewline
\textit{\small{}{}{}polar bear}\tabularnewline
\end{tabular}{\small{}{}{}{ }}\tabularnewline
\midrule
{\small{}{}{}{Raccoon }}  & {\small{}{}{}{}}%
\begin{tabular}{@{}l@{}}
{\small{}{}{}{0.4117}}\tabularnewline
{\small{}{}{}{skunk}}\tabularnewline
\end{tabular} & {\small{}{}{}{}}%
\begin{tabular}{@{}l@{}}
{\small{}{}{}{0.3385}}\tabularnewline
{\small{}{}{}{wolf}}\tabularnewline
\end{tabular} & {\small{}{}{}{}}%
\begin{tabular}{@{}l@{}}
{\small{}{}{}{0.3035}}\tabularnewline
{\small{}{}{}{squirrel}}\tabularnewline
\end{tabular} & \multicolumn{1}{l}{{\small{}{}{}{}}%
\begin{tabular}{@{}l@{}}
\textit{\small{}{}{}-0.0202}\tabularnewline
\textit{\small{}{}{}lion}\tabularnewline
\end{tabular}} & \multicolumn{1}{l}{{\small{}{}{}{}}%
\begin{tabular}{@{}l@{}}
\textit{\small{}{}{}-0.0155}\tabularnewline
\textit{\small{}{}{}horse}\tabularnewline
\end{tabular}} & {\small{}{}{}{}}%
\begin{tabular}{@{}l@{}}
\textit{\small{}{}{}0.0142}\tabularnewline
\textit{\small{}{}{}killer whale}\tabularnewline
\end{tabular}{\small{}{}{}{ }}\tabularnewline
\bottomrule
\end{tabular}{\small{}\caption{{{Regression weights of three target domain (unseen) animals on
AwA.}}}
\label{tab:weights}}
\end{table*}

\subsubsection{Visualization and Interpretation}

\label{sec:exp-vis}In this section, we visualize the strong and weak
signals learned in the zero-shot learning tasks of the embedding model.
In particular, we utilize the AwA dataset as the testbed. Among all
the 50 coarse-grained animals, 10 target classes are regressed by
40 auxiliary source classes with corresponding weights. In other
words, each target class can be represented as the linear combination
of existing 40 source classes. We visualize the linear regression
weights on three target classes in Tab. \ref{tab:weights}.  {
We sort the absolute value of weights of strong signals ($\tilde{\beta}$) 
and weak signals (comparably large value in $\vert \beta-\tilde{\beta} \vert$). Then we display the largest 3 strong and weak signals respectively.}

\noindent \textbf{Strong Signals.} The strong signals  {(with large magnitudes)} imply the strong
correlations between the target animals and source animals. This can
be clearly showed by the weights of strong signals. For example, \textquotedbl{}cow\textquotedbl{},
\textquotedbl{}rhinoceros\textquotedbl{}, and \textquotedbl{}ox\textquotedbl{}
have similar shape (hooves, tail), size (big) as \textquotedbl{}pig\textquotedbl{};
and thus the weights of these strong signals, are 0.4267 (cow), 0.1837
(rhinoceros) and 0.1375 (ox) individually. These strong signals are
well captured by our model.

\noindent \textbf{Weak Signals.} The weak signals  {(with small magnitudes)} indicate the
relatively  {weak} correlation between the source animals and the target
animals. For instance, \textquotedbl{}hamster\textquotedbl{}, \textquotedbl{}skunk\textquotedbl{}
have very different visual appearance from \textquotedbl{}pig\textquotedbl{},
while their only possible relationship may be the similar habitats.
Thanks to the decomposition ability of our {MSplit} LBI model,
 {these weak signals can be captured to further help learn the
embedding, and hence our method can achieve better classification result (see Tab. ~\ref{tab:comparison}).}

\subsection{Few-shot Learning}

\noindent \textbf{Datasets.} We test our method on two datasets, namely Omniglot
\citet{lake2011one} and SUN attribute dataset (SUN) \cite{forinaextendible}. Omniglot is a handwriting dataset with 1,623
characters from 50 alphabets. Each character has 20 handwriting images. There are 14,340 images belonging to 717 classes in SUN. 102 attributes are
annotated for all images.

\begin{table}
\centering{}{\small{}\label{my-label} }%
\small
\begin{tabular}{cccccc}
\hline
\multirow{2}{*}{{\small{}Method }} & \multirow{2}{*}{{\small{}Finetune}} & \multicolumn{2}{c}{{\small{}{}5-way}} & \multicolumn{2}{c}{{\small{}{}20-way}}\tabularnewline
\cline{3-6}
 &  & \multicolumn{1}{c}{{\small{}{}1-shot }} & \multicolumn{1}{c}{{\small{}{}5-shot }} & \multicolumn{1}{c}{{\small{}{}1-shot }} & {\small{}{}5-shot }\tabularnewline
\hline
 \hline
{\small{}{}{}MANN}  & {\small{}{}{}N}  & {\small{}{}{}{82.8}}  & {\small{}{}{}{94.9}}  & {\small{}{}{}{-}}  & {\small{}{}{}{-} }\tabularnewline
\hline
\multirow{2}{*}{{\small{}C-Siam} } & \multicolumn{1}{c}{{\small{}{}{}N }} & \multicolumn{1}{c}{{\small{}{}{}96.7 }} & \multicolumn{1}{c}{{\small{}{}{}98.4 }} & \multicolumn{1}{c}{{\small{}{}{}88.0 }} & {\small{}{}{}96.5 }\tabularnewline
\cline{2-6}
 & {\small{}{}{}Y}  & {\small{}{}{}97.3}  & {\small{}{}{}98.4}  & {\small{}{}{}88.1}  & {\small{}{}{}97.0 }\tabularnewline
\hline
\multirow{2}{*}{{\small{}M-Net} } & {\small{}{}{}N}  & \textbf{\small{}{}{}98.1}  & {\small{}{}{}98.9}  & \textbf{\small{}{}{}93.8}  & {\small{}{}{}98.5 }\tabularnewline
\cline{2-6}
 & {\small{}{}{}Y}  & {\small{}{}{}97.9}  & {\small{}{}{}98.7}  & {\small{}{}{}93.5}  & \textbf{\small{}{}{}98.7 }\tabularnewline
\hline
\hline
{\small{}{}{}Lasso}  & {\small{}{}{}N}  & {\small{}{}{}94.7}  & {\small{}{}{}99.1}  & {\small{}{}{}85.3}  & {\small{}{}{}97.4 }\tabularnewline
\hline
{\small{}{}{}Ridge}  & {\small{}{}{}N}  & {\small{}{}{}96.8}  & \textbf{\small{}{}{}99.4}  & {\small{}{}{}88.7}  & {\small{}{}{}97.5 }\tabularnewline
\hline
{\small{}{}{}Ours ($\tilde{\beta}$)}  & {\small{}{}{}N}  & {\small{}{}{}94.7} & {\small{}{}{}98.9}  &  {\small{}{}{}83.7} & {\small{}{}{}97.3 }\tabularnewline
\hline
{\small{}{}{}Ours (${\beta}$)}  & {\small{}{}{}N}  & {\small{}{}{}94.8}  & {\small{}{}{}99.2}  & {\small{}{}{}83.7} & {\small{}{}{}97.6 }\tabularnewline
\hline
\end{tabular}{\small{}\caption{{\small{}{}{}\label{tab:Results-on-few-shot}
{Few-shot learning performance on Omniglot dataset.}}}
}
\end{table}

\begin{table}[]
\centering{}{\small{}\label{my-label} }%
\small
\begin{tabular}{c|c|c|c}
\hline
Method       & LASSO & Ours ($\tilde{\beta}$)  & Ours ($\beta$) \\ \hline
Accuracy(\%) & 59.09 & 59.32 & \textbf{61.47}                          \\ \hline
\end{tabular}
\caption{{Few-shot learning performance on SUN dataset.} }
\label{tab:sun}
\end{table}

\noindent \textbf{Setting-Omniglot.}  { We implement the basic few-shot learning task, \emph{i.e.}
$N$ way $k$-shot learning task. We have $k$ labeled training samples from
each of $N$ target domain classes. The rest instances from these
$N$ classes are utilized as testing data  {(chance level $= 1/N$).}
For Omniglot, we  follow the setting in \citet{matchingnet_1shot} in which 1,200 characters
are used for source domain, while the rest are the target domain.
We choose the MobileNet \citet{howard2017mobilenets} as the feature extractor,
then train it on the source domain. The data augmentation strategy
including rotation and shift is the same as that in \citet{matchingnet_1shot}.
The model is trained via SGD optimizer with learning rate 0.05. Then
the trained model is utilized to extract features for the target domain.
For speeding up the experiments, we further use PCA \cite{bishop1999vpca} to realize dimensionality reduction and obtain 40-dim features.
We compare all methods under four settings: 5 way 1-shot/5-shot
and 20-way 1-shot/5-shot.
}

\noindent \textbf{Setting-SUN.} {For SUN dataset, we consider all classes as the target domain. The 102-dim attributes are
utilized as the features. We implement 5 way 1-shot image classification.}

\noindent \textbf{Result-Omniglot.} {In Tab. \ref{tab:Results-on-few-shot}, we compare our method with
several baselines, including MANN \citet{santoro2016meta}, C-Siam (Convolutional Siamese
Net \citet{siamese_1shot}), M-Net (Matching Networks \citet{matchingnet_1shot}),
Ridge (Ridge Regression) and Lasso on Omniglot dataset. It shows that our results outperform Lasso in most settings, except 20-way 1-shot setting. On this dataset, Ridge performs better than ours. One possible reason is that, after dimensionality reduction, most signals are strong ones. Hence, further feature selection may damage the performance. The deep model M-Net achieves the best performance in most settings. We further highlight several observations.}

\noindent (1) \emph{The gap between our method and M-Net narrows in 5-shot settings.}
A possible reason is that when only one sample is provided, the calculating of linear embedding is not stable. This phenomenon is also viewed in Lasso and Ridge regression.

\noindent (2) \emph{Doing the fine-tuning may not matter.}
The improvement (averagely 0.4\%) brought by fine-tune is slight in C-Siam. In contrast, doing fine-tune in M-Net depresses the performance in most settings.

\noindent {(3) }{\emph{Linear models can achieve state-of-the-art performance.} These linear models (Lasso, Ridge and MSplit LBI) achieves state-of-the-art classification accuracies in few-shot learning, compared to deep models.  }

\textbf{Result-SUN.} {Tab. \ref{tab:sun} shows the 5 way 1 shot image classification results of LASSO and our method on SUN dataset. The classification accuracies for Lasso, ours($\tilde{\beta}$) and ours($\beta$) are 59.09\%, 59.32\% and 61.47\% respectively, which means ours($\beta$) outperforms ours($\tilde{\beta}$) by 2.15\% due to additional weak signals.}

\section{Conclusion \& Future Work}

In this paper, we assume that the features consist of sparse strong signals, dense weak signals and random
noise.
Hence, we propose the novel MSplit LBI to capture both strong and
weak signals. Our method can realize the feature selection (i.e. capture
strong signals) and dense estimation (i.e. additionally capture weak signals) simultaneously.
We prove both theoretical and experimental comparison to the $L_{1}$
(lasso) and $L_{2}$ (ridge) regularization terms which show advantages
of our method. Experiments on simulation data and four popular datasets
in few-shot and zero-shot learning show that our method achieves state-of-the-art
performance.

As our MSplit LBI is a kind of regularization method, it can be integrated
in many regression/classification models. A natural future work is
the integration of MSplit LBI and deep neural networks, which may
split sparse strong signals and dense weak signals at the headstream
of feature extraction.

\newpage
\section*{Acknowledgement}
This work was supported in part by the following grants 973-2015CB351800, NSFC-61625201, NSFC-61527804, NSFC-61702108 and Eastern Scholar (TP2017006).

{\small{}{}{}  \bibliographystyle{plainnat}
\bibliography{example_paper}

\begin{thebibliography}{57}
\providecommand{\natexlab}[1]{#1}
\providecommand{\url}[1]{\texttt{#1}}
\expandafter\ifx\csname urlstyle\endcsname\relax
  \providecommand{\doi}[1]{doi: #1}\else
  \providecommand{\doi}{doi: \begingroup \urlstyle{rm}\Url}\fi

\bibitem[Akata et~al.(2013)Akata, Perronnin, Harchaoui, and
  Schmid]{akata2013label}
Zeynep Akata, Florent Perronnin, Zaid Harchaoui, and Cordelia Schmid.
\newblock Label-embedding for attribute-based classification.
\newblock In \emph{Proceedings of the IEEE Conference on Computer Vision and
  Pattern Recognition}, pages 819--826, 2013.

\bibitem[Akata et~al.(2015)Akata, Reed, Walter, Lee, and
  Schiele]{akata2015evaluation}
Zeynep Akata, Scott Reed, Daniel Walter, Honglak Lee, and Bernt Schiele.
\newblock Evaluation of output embeddings for fine-grained image
  classification.
\newblock In \emph{CVPR}, 2015.

\bibitem[Ba et~al.(2015)Ba, Swersky, Fidler, and Salakhutdinov]{deep_0shot}
Jimmy~Lei Ba, Kevin Swersky, Sanja Fidler, and Ruslan Salakhutdinov.
\newblock Predicting deep zero-shot convolutional neural networks using textual
  descriptions.
\newblock In \emph{ICCV}, 2015.

\bibitem[Bishop(1999)]{bishop1999vpca}
Christopher~M. Bishop.
\newblock Variational principal components.
\newblock 1999.

\bibitem[Boyd et~al.(2011)Boyd, Parikh, Chu, Peleato, Eckstein,
  et~al.]{boyd2011distributed}
Stephen Boyd, Neal Parikh, Eric Chu, Borja Peleato, Jonathan Eckstein, et~al.
\newblock Distributed optimization and statistical learning via the alternating
  direction method of multipliers.
\newblock \emph{Foundations and Trends{\textregistered} in Machine Learning},
  3\penalty0 (1):\penalty0 1--122, 2011.

\bibitem[Changpinyo et~al.(2016)Changpinyo, Chao, Gong, and
  Sha]{changpinyo2016synthesized}
Soravit Changpinyo, Wei-Lun Chao, Boqing Gong, and Fei Sha.
\newblock Synthesized classifiers for zero-shot learning.
\newblock In \emph{CVPR}, pages 5327--5336, 2016.

\bibitem[Deng et~al.(2009)Deng, Dong, Socher, Li, Li, and
  Fei-Fei]{deng2009imagenet}
Jia Deng, Wei Dong, Richard Socher, Li-Jia Li, Kai Li, and Li~Fei-Fei.
\newblock Imagenet: A large-scale hierarchical image database.
\newblock In \emph{CVPR}, pages 248--255, 2009.

\bibitem[Ding et~al.(2017)Ding, Shao, and Fu]{dinglow}
Zhengming Ding, Ming Shao, and Yun Fu.
\newblock Low-rank embedded ensemble semantic dictionary for zero-shot
  learning.
\newblock \emph{CVPR}, 2017.

\bibitem[Elhamifar and Vidal(2009)]{elhamifar2009sparse}
Ehsan Elhamifar and Ren{\'e} Vidal.
\newblock Sparse subspace clustering.
\newblock In \emph{CVPR}, pages 2790--2797, 2009.

\bibitem[Fan and Li(2001)]{FanLi01}
Jianqing Fan and Runze Li.
\newblock Variable selection via nonconcave penalized likelihood and its oracle
  properties.
\newblock 2001.

\bibitem[Fei-Fei et~al.(2006)Fei-Fei, Fergus, and Perona]{fei2006one}
Li~Fei-Fei, Rob Fergus, and Pietro Perona.
\newblock One-shot learning of object categories.
\newblock \emph{IEEE transactions on pattern analysis and machine
  intelligence}, 28\penalty0 (4):\penalty0 594--611, 2006.

\bibitem[Forina()]{forinaextendible}
M~Forina.
\newblock An extendible package for data exploration.
\newblock \emph{Classification and Correlation. Institute of Pharmaceutical and
  Food Analysis and Technologies, Genoa, Italy}.

\bibitem[Forman(2003)]{forman2003extensive}
George Forman.
\newblock An extensive empirical study of feature selection metrics for text
  classification.
\newblock \emph{Journal of machine learning research}, 3\penalty0
  (Mar):\penalty0 1289--1305, 2003.

\bibitem[Fu and Sigal(2016)]{fu2016semi}
Yanwei Fu and Leonid Sigal.
\newblock Semi-supervised vocabulary-informed learning.
\newblock In \emph{CVPR}, pages 5337--5346, 2016.

\bibitem[Fu et~al.(2015{\natexlab{a}})Fu, Hospedales, Xiang, and
  Gong]{fu2015transductive}
Yanwei Fu, Timothy~M Hospedales, Tao Xiang, and Shaogang Gong.
\newblock Transductive multi-view zero-shot learning.
\newblock \emph{TPAMI}, 37\penalty0 (11):\penalty0 2332--2345,
  2015{\natexlab{a}}.

\bibitem[Fu et~al.(2015{\natexlab{b}})Fu, Hospedales, Xiang, and
  Gong]{transductiveEmbeddingJournal}
Yanwei Fu, Timothy~M. Hospedales, Tao Xiang, and Shaogang Gong.
\newblock Transductive multi-view zero-shot learning.
\newblock \emph{IEEE TPAMI}, 2015{\natexlab{b}}.

\bibitem[Guo et~al.(2017)Guo, Ding, Han, and Gao]{guo2017zero}
Yuchen Guo, Guiguang Ding, Jungong Han, and Yue Gao.
\newblock Zero-shot recognition via direct classifier learning with transferred
  samples and pseudo labels.
\newblock In \emph{AAAI}, 2017.

\bibitem[He et~al.(2016)He, Zhang, Ren, and Sun]{he2016deep}
Kaiming He, Xiangyu Zhang, Shaoqing Ren, and Jian Sun.
\newblock Deep residual learning for image recognition.
\newblock In \emph{CVPR}, pages 770--778, 2016.

\bibitem[Howard et~al.(2017)Howard, Zhu, Chen, Kalenichenko, Wang, Weyand,
  Andreetto, and Adam]{howard2017mobilenets}
Andrew~G. Howard, Menglong Zhu, Bo~Chen, Dmitry Kalenichenko, Weijun Wang,
  Tobias Weyand, Marco Andreetto, and Hartwig Adam.
\newblock Mobilenets: Efficient convolutional neural networks for mobile vision
  applications.
\newblock In \emph{arxiv}, 2017.

\bibitem[Huang et~al.(2016)Huang, Sun, Xiong, and Yao]{Splitlbi}
Chendi Huang, Xinwei Sun, Jiechao Xiong, and Yuan Yao.
\newblock Split lbi: An iterative regularization path with structural sparsity.
  advances in neural information processing systems.
\newblock \emph{Advances In Neural Information Processing Systems}, pages
  3369--3377, 2016.

\bibitem[Kabir et~al.(2010)Kabir, Islam, and Murase]{kabir2010new}
Md~Monirul Kabir, Md~Monirul Islam, and Kazuyuki Murase.
\newblock A new wrapper feature selection approach using neural network.
\newblock \emph{Neurocomputing}, 73\penalty0 (16-18):\penalty0 3273--3283,
  2010.

\bibitem[Kavukcuoglu et~al.(2010)Kavukcuoglu, Ranzato, and
  LeCun]{kavukcuoglu2010fast}
Koray Kavukcuoglu, Marc'Aurelio Ranzato, and Yann LeCun.
\newblock Fast inference in sparse coding algorithms with applications to
  object recognition.
\newblock \emph{arXiv preprint arXiv:1010.3467}, 2010.

\bibitem[Koch et~al.(2015{\natexlab{a}})Koch, Zemel, and
  Salakhutdinov]{koch2015siamese}
Gregory Koch, Richard Zemel, and Ruslan Salakhutdinov.
\newblock Siamese neural networks for one-shot image recognition.
\newblock In \emph{ICML Deep Learning Workshop}, volume~2, 2015{\natexlab{a}}.

\bibitem[Koch et~al.(2015{\natexlab{b}})Koch, Zemel, and
  Salakhutdinov]{siamese_1shot}
Gregory Koch, Richard Zemel, and Ruslan Salakhutdinov.
\newblock Siamese neural networks for one-shot image recognition.
\newblock In \emph{ICML -- Deep Learning Workshok}, 2015{\natexlab{b}}.

\bibitem[Kodirov et~al.(2015)Kodirov, Xiang, Fu, and
  Gong]{kodirov2015unsupervised}
Elyor Kodirov, Tao Xiang, Zhenyong Fu, and Shaogang Gong.
\newblock Unsupervised domain adaptation for zero-shot learning.
\newblock In \emph{ICCV}, pages 2452--2460, 2015.

\bibitem[Krizhevsky et~al.(2012)Krizhevsky, Sutskever, and
  Hinton]{krizhevsky2012imagenet}
Alex Krizhevsky, Ilya Sutskever, and Geoffrey~E Hinton.
\newblock Imagenet classification with deep convolutional neural networks.
\newblock In \emph{NIPS}, pages 1097--1105, 2012.

\bibitem[Lake et~al.(2011)Lake, Salakhutdinov, Gross, and
  Tenenbaum]{lake2011one}
Brenden Lake, Ruslan Salakhutdinov, Jason Gross, and Joshua Tenenbaum.
\newblock One shot learning of simple visual concepts.
\newblock In \emph{Proceedings of the Annual Meeting of the Cognitive Science
  Society}, volume~33, 2011.

\bibitem[Lampert et~al.(2013)Lampert, Nickisch, and
  Harmeling]{lampert13AwAPAMI}
Christoph~H. Lampert, Hannes Nickisch, and Stefan Harmeling.
\newblock Attribute-based classification for zero-shot visual object
  categorization.
\newblock \emph{IEEE TPAMI}, 2013.

\bibitem[Lampert et~al.(2014)Lampert, Nickisch, and
  Harmeling]{lampert2014attribute}
Christoph~H Lampert, Hannes Nickisch, and Stefan Harmeling.
\newblock Attribute-based classification for zero-shot visual object
  categorization.
\newblock \emph{TPAMI}, 36\penalty0 (3):\penalty0 453--465, 2014.

\bibitem[Lee et~al.(2015)Lee, Sun, Liu, and Taylor]{lee2015communication}
Jason~D Lee, Yuekai Sun, Qiang Liu, and Jonathan~E Taylor.
\newblock Communication-efficient sparse regression: a one-shot approach.
\newblock \emph{arXiv preprint arXiv:1503.04337}, 2015.

\bibitem[Li et~al.(2015)Li, Guo, and Schuurmans]{li2015semi}
Xin Li, Yuhong Guo, and Dale Schuurmans.
\newblock Semi-supervised zero-shot classification with label representation
  learning.
\newblock In \emph{ICCV}, pages 4211--4219, 2015.

\bibitem[Mairal et~al.(2009)Mairal, Bach, Ponce, Sapiro, and
  Zisserman]{mairal2009non}
Julien Mairal, Francis Bach, Jean Ponce, Guillermo Sapiro, and Andrew
  Zisserman.
\newblock Non-local sparse models for image restoration.
\newblock In \emph{Computer Vision, 2009 IEEE 12th International Conference
  on}, pages 2272--2279. IEEE, 2009.

\bibitem[Osher et~al.(2016)Osher, Ruan, Xiong, Yao, and Yin]{bregman}
Stanley Osher, Feng Ruan, Jiechao Xiong, Yuan Yao, and Wotao Yin.
\newblock Sparse recovery via differential inclusions.
\newblock \emph{Applied and Computational Harmonic Analysis}, 2016.

\bibitem[Palatucci et~al.(2009)Palatucci, Pomerleau, Hinton, and
  Mitchell]{palatucci2009zero}
Mark Palatucci, Dean Pomerleau, Geoffrey~E Hinton, and Tom~M Mitchell.
\newblock Zero-shot learning with semantic output codes.
\newblock In \emph{NIPS}, pages 1410--1418, 2009.

\bibitem[Romera-Paredes and Torr(2015)]{romera2015embarrassingly}
Bernardino Romera-Paredes and PHS Torr.
\newblock An embarrassingly simple approach to zero-shot learning.
\newblock In \emph{ICML}, 2015.

\bibitem[Saeys et~al.(2007)Saeys, Inza, and Larra{\~n}aga]{saeys2007review}
Yvan Saeys, I{\~n}aki Inza, and Pedro Larra{\~n}aga.
\newblock A review of feature selection techniques in bioinformatics.
\newblock \emph{bioinformatics}, 23\penalty0 (19):\penalty0 2507--2517, 2007.

\bibitem[Santoro et~al.(2016)Santoro, Bartunov, Botvinick, Wierstra, and
  Lillicrap]{santoro2016meta}
Adam Santoro, Sergey Bartunov, Matthew Botvinick, Daan Wierstra, and Timothy
  Lillicrap.
\newblock Meta-learning with memory-augmented neural networks.
\newblock In \emph{International conference on machine learning}, pages
  1842--1850, 2016.

\bibitem[Simonyan and Zisserman(2014)]{simonyan2014very}
Karen Simonyan and Andrew Zisserman.
\newblock Very deep convolutional networks for large-scale image recognition.
\newblock \emph{arXiv preprint arXiv:1409.1556}, 2014.

\bibitem[Snell et~al.(2017)Snell, Swersky, and Zemel]{snell2017prototypical}
Jake Snell, Kevin Swersky, and Richard~S Zemel.
\newblock Prototypical networks for few-shot learning.
\newblock \emph{arXiv preprint arXiv:1703.05175}, 2017.

\bibitem[Sun et~al.(2017)Sun, Hu, Yao, and Wang]{sun2017gsplit}
Xinwei Sun, Lingjing Hu, Yuan Yao, and Yizhou Wang.
\newblock Gsplit lbi: Taming the procedural bias in neuroimaging for disease
  prediction.
\newblock In \emph{International Conference on Medical Image Computing and
  Computer-Assisted Intervention}, pages 107--115. Springer, 2017.

\bibitem[Szegedy et~al.(2016)Szegedy, Ioffe, and Vanhoucke]{InceptionNet}
Christian Szegedy, Sergey Ioffe, and Vincent Vanhoucke.
\newblock Inception-v4, inception-resnet and the impact of residual connections
  on learning.
\newblock \emph{CoRR}, abs/1602.07261, 2016.
\newblock URL \url{http://arxiv.org/abs/1602.07261}.

\bibitem[Vinyals et~al.(2016)Vinyals, Blundell, Lillicrap, Kavukcuoglu, and
  Wierstra]{matchingnet_1shot}
Oriol Vinyals, Charles Blundell, Timothy Lillicrap, Koray Kavukcuoglu, and Daan
  Wierstra.
\newblock Matching networks for one shot learning.
\newblock In \emph{NIPS}, 2016.

\bibitem[Wah et~al.(2011)Wah, Branson, Welinder, Perona, and
  Belongie]{WahCUB_200_2011}
C.~Wah, S.~Branson, P.~Welinder, P.~Perona, and S.~Belongie.
\newblock {The Caltech-UCSD Birds-200-2011 Dataset}.
\newblock Technical report, California Institute of Technology, 2011.

\bibitem[Wahlberg et~al.(2012)Wahlberg, Boyd, Annergren, and
  Wang]{wahlberg2012admm}
Bo~Wahlberg, Stephen Boyd, Mariette Annergren, and Yang Wang.
\newblock An admm algorithm for a class of total variation regularized
  estimation problems.
\newblock \emph{IFAC Proceedings Volumes}, 45\penalty0 (16):\penalty0 83--88,
  2012.

\bibitem[Wainwright(2009)]{sharp_lasso}
Martin~J. Wainwright.
\newblock Sharp thresholds for high-dimensional and noisy sparsity recovery
  using l1-constrained quadratic programming (lasso).
\newblock \emph{IEEE Transactions on Information Theory}, 2009.

\bibitem[Wang et~al.(2016)Wang, Li, Lin, and Zhuang]{wang2016relational}
Donghui Wang, Yanan Li, Yuetan Lin, and Yueting Zhuang.
\newblock Relational knowledge transfer for zero-shot learning.
\newblock In \emph{Thirtieth AAAI}, 2016.

\bibitem[Wright et~al.(2009)Wright, Yang, Ganesh, Sastry, and
  Ma]{wright2009robust}
John Wright, Allen~Y Yang, Arvind Ganesh, S~Shankar Sastry, and Yi~Ma.
\newblock Robust face recognition via sparse representation.
\newblock \emph{IEEE transactions on pattern analysis and machine
  intelligence}, 31\penalty0 (2):\penalty0 210--227, 2009.

\bibitem[Xian et~al.(2016)Xian, Akata, Sharma, Nguyen, Hein, and
  Schiele]{xian2016latent}
Yongqin Xian, Zeynep Akata, Gaurav Sharma, Quynh Nguyen, Matthias Hein, and
  Bernt Schiele.
\newblock Latent embeddings for zero-shot classification.
\newblock In \emph{CVPR}, pages 69--77, 2016.

\bibitem[Xu et~al.(2017)Xu, Shen, Yang, Zhang, Shen, and Song]{xu2017matrix}
Xing Xu, Fumin Shen, Yang Yang, Dongxiang Zhang, Heng~Tao Shen, and Jingkuan
  Song.
\newblock Matrix tri-factorization with manifold regularizations for zero-shot
  learning.
\newblock In \emph{CVPR}, 2017.

\bibitem[Ye and Xie(2011)]{ye2011split}
Gui-Bo Ye and Xiaohui Xie.
\newblock Split bregman method for large scale fused lasso.
\newblock \emph{Computational Statistics \& Data Analysis}, 55\penalty0
  (4):\penalty0 1552--1569, 2011.

\bibitem[Yu and Liu(2003)]{yu2003feature}
Lei Yu and Huan Liu.
\newblock Feature selection for high-dimensional data: A fast correlation-based
  filter solution.
\newblock In \emph{Proceedings of the 20th international conference on machine
  learning (ICML-03)}, pages 856--863, 2003.

\bibitem[Zhang et~al.(2017)Zhang, Xiang, and Gong]{zhang2016learning}
Li~Zhang, Tao Xiang, and Shaogang Gong.
\newblock Learning a deep embedding model for zero-shot learning.
\newblock \emph{CVPR}, 2017.

\bibitem[Zhang and Saligrama(2016)]{zhang2016zero}
Ziming Zhang and Venkatesh Saligrama.
\newblock Zero-shot learning via joint latent similarity embedding.
\newblock In \emph{CVPR}, 2016.

\bibitem[Zhao et~al.(2017)Zhao, Wu, Wu, and Wang]{zhao2017zero}
Bo~Zhao, Botong Wu, Tianfu Wu, and Yizhou Wang.
\newblock Zero-shot learning posed as a missing data problem.
\newblock In \emph{Proceedings of ICCV Workshops}, pages 2616--2622, 2017.

\bibitem[Zhao and Yu(2006{\natexlab{a}})]{modelselection_jmlr}
Peng Zhao and Bin Yu.
\newblock On model selection consistency of lasso.
\newblock \emph{JMLR}, 2006{\natexlab{a}}.

\bibitem[Zhao and Yu(2006{\natexlab{b}})]{zhao2006model}
Peng Zhao and Bin Yu.
\newblock On model selection consistency of lasso.
\newblock \emph{Journal of Machine learning research}, 7\penalty0
  (Nov):\penalty0 2541--2563, 2006{\natexlab{b}}.

\bibitem[Zou and Hastie(2005)]{zou2005regularization}
Hui Zou and Trevor Hastie.
\newblock Regularization and variable selection via the elastic net.
\newblock \emph{Journal of the Royal Statistical Society: Series B (Statistical
  Methodology)}, 67\penalty0 (2):\penalty0 301--320, 2005.

\end{thebibliography}
 } \bibliographystyle{icml2018}


\appendix
\end{document}